\documentclass{article}

\PassOptionsToPackage{numbers, compress}{natbib}
\usepackage[final]{nips_2017}

\usepackage[utf8]{inputenc} 
\usepackage[T1]{fontenc}    
\usepackage{hyperref}       
\usepackage{url}            
\usepackage{booktabs}       
\usepackage{amsfonts}       
\usepackage{nicefrac}       
\usepackage{microtype}      
\usepackage{graphicx}		

\title{Detection-aided liver lesion segmentation \\using deep learning}

\author{
 M\'{i}riam Bellver\raisebox{5pt}{\small \dag}, Kevis-Kokitsi Maninis\raisebox{5pt}{\small \ddag}, Jordi Pont-Tuset\raisebox{5pt}{\small \ddag}, Xavier Gir{\'o}-i-Nieto\raisebox{5pt}{\small \S}, \\
  \textbf{Jordi Torres \raisebox{5pt}{\small \dag,\S}, Luc Van Gool \raisebox{5pt}{\small \ddag} }\\
  \raisebox{5pt}{\small \dag}Barcelona Supercomputing Center (BSC) \\
  \raisebox{5pt}{\small \ddag} Eidgen\"{o}ssische Technische Hochschule Z\"{u}rich (ETH Z\"{u}rich) \\
  \raisebox{5pt}{\small \S}Universitat Polit{\`e}cnica de Catalunya (UPC) \\
}

\begin{document}

\maketitle

\begin{abstract}
  A fully automatic technique for segmenting the liver and localizing its unhealthy tissues is a convenient tool in order to diagnose hepatic diseases and assess the response to the according treatments. In this work we propose a method to segment the liver and its lesions from Computed Tomography (CT) scans using Convolutional Neural Networks (CNNs), that have proven good results in a variety of computer vision tasks, including medical imaging. The network that segments the lesions consists of a cascaded architecture, which first focuses on the region of the liver in order to segment the lesions on it. Moreover, we train a detector to localize the lesions, and mask the results of the segmentation network with the positive detections. The segmentation architecture is based on DRIU~\cite{maninis2016deep}, a Fully Convolutional Network (FCN) with side outputs that work on feature maps of different resolutions, to finally  benefit from the multi-scale information learned by different stages of the network. The main contribution of this work is the use of a detector to localize the lesions, which we show to be beneficial to remove false positives triggered by the segmentation network. Source code and models are available
at \url{https://imatge-upc.github.io/liverseg-2017-nipsws/}.
\end{abstract}

\section{Introduction}

Segmenting the liver and its lesions on medical images helps oncologists to accurately diagnose liver cancer, as well as to assess the treatment response of patients. Typically, doctors rely on manual segmentation techniques in order to interpret the Contrast Tomography (CT) and Magnetic Resonance Imaging (MRI) images. Automatic tools that are not as subjective and time-consuming have been widely studied in the recent years. Liver lesion segmentation is a challenging task due to the low contrast between liver, lesions, and also nearby organs. Other additional difficulties are the lesion size variability and the noise in CT scans. Building a robust system that is able to beat these difficulties is still an open problem. Recently, methods based on deep Convolutional Neural Networks (CNNs) have demonstrated to be robust to these challenges, and have achieved the state of the art at this task \cite{christ2017automatic, han2017automatic, bi2017automatic}. 

In this paper we adapt DRIU~\cite{maninis2016deep} for the task of segmenting both the liver and its lesions from CT scans. DRIU is a Fully Convolutional Network (FCN) that has side outputs with supervision at different convolutional stages. This architecture has proven to be successful for the medical task of segmenting the blood vessels and optical disk of eye fundus images, as well as for video object segmentation~\cite{caelles2016one} in generic videos. The core of our network for lesion and liver segmentation consists in using the strength of a segmentation network plus a detection network to localize the lesions. For training all the networks we used the Liver Tumor Segmentation (LiTS) dataset, which is composed of 131 CT scans for training and 70 for testing. 

\section{Detection-aided liver and its lesions segmentation using deep learning}

\begin{figure*}[h!]
  \centering
  \includegraphics[width=0.80\textwidth]{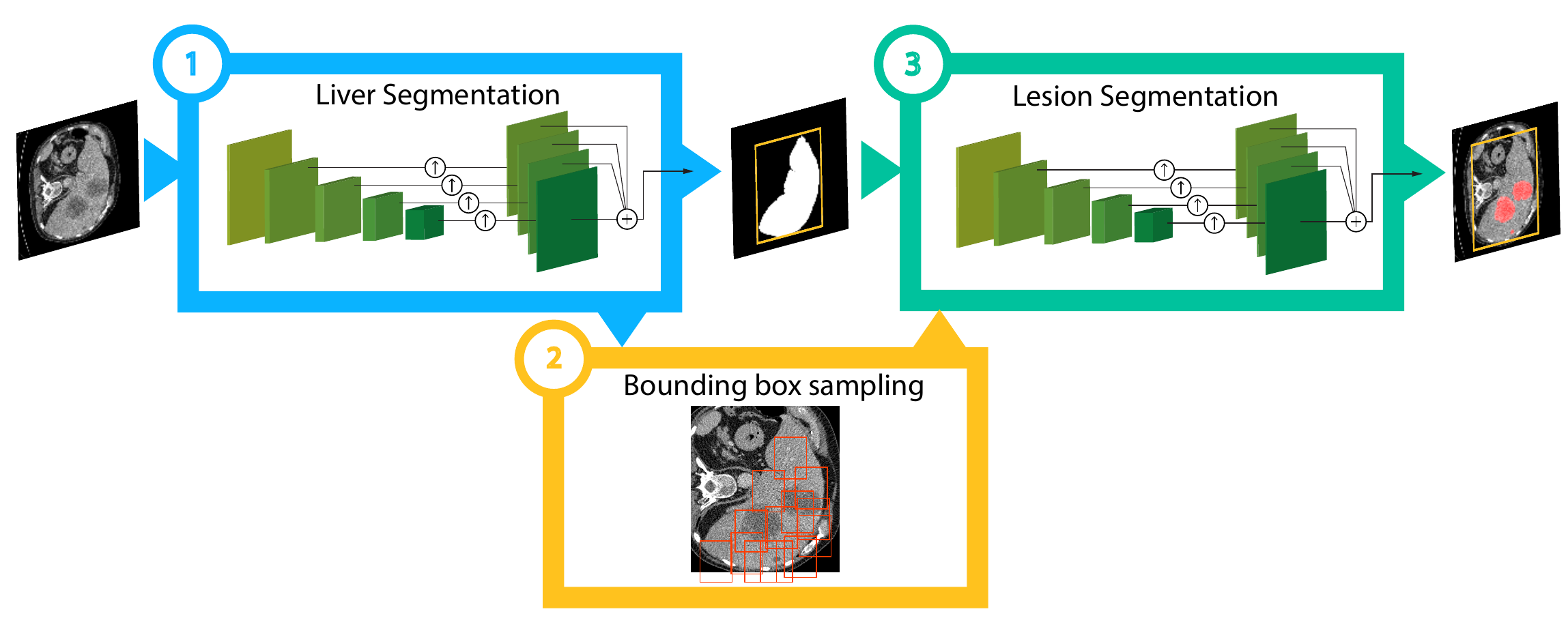}
  \caption[Stack of slices as input]{Architecture for the detection-aided liver and its lesions segmentation. The first stage consists in segmenting the liver. Once we have the liver prediction, we place a 3D bounding box around the liver, and the different slices cropped by this bounding box are segmented by the lesion segmentation network. The detection network operates on patches sampled around the liver. Finally, we only keep the positive detections that agree with the segmentation of the lesion.}
  \label{Image:architecture}
\end{figure*}

Our pipeline is illustrated in Figure~\ref{Image:architecture}. It is a cascaded architecture, which first segments the liver to focus on the region of interest in order to segment the lesion. In this section, we first present the baseline segmentation architecture, and then the different features implemented to adapt it to the liver and lesion segmentation task.

\subsection{Segmentation network}
\label{seg-net}

The lesion and liver segmentation networks are both based on DRIU~\cite{maninis2016deep}, an architecture for retinal image segmentation that segments the blood vessels and optic disc on fundus images. The architecture uses VGG-16~\cite{simonyan2014very} as the base network, removing the last fully connected layers, so that the network consists of convolutional, ReLU, and max-pooling layers. The base network is pre-trained on Imagenet~\cite{ILSVRC15} and consists of a set of convolutional stages, each of them working at the same feature map resolution, separated by the pooling layers. As the network goes deeper, the information is coarser and  the learned features are more related to semantics. On the other hand, at the shallower feature maps that work at a higher resolution, filters capture more local information. To take advantage of the information learned at feature maps that work at different resolutions, DRIU uses several side outputs with supervision. A side output is a set of convolutional layers that are connected at the end of a specific convolutional stage from the base network. Each of these side outputs specializes on different types of features, depending on the resolution at the connection point. The feature maps produced by each side output are resized and linearly combined to output the final result. 

\subsubsection{Loss objective}

Regarding the loss objective for training the segmentation network, we worked with a weighted version of the Binary Cross Entropy (BCE) loss due to the imbalance between positive and negative lesion pixels in the database. The weighted BCE is defined as:
\[\mathit{L}(y,\hat{y}) = - (1-w)*y\log \hat{y} - w*(1-y)\log (1-\hat{y})\]
where \(\hat{y}\) is the predicted mask and \(y\) is the ground truth.

For training each segmentation network we compute a weighting term for the foreground and another for the background class. The weighting term for the foreground class is computed by summing all the pixels belonging to the foreground divided by total number of pixels. It is important to notice that only the slices that contain the foreground class are considered, as proposed in~\cite{eigen2015predicting}. The same is done for the background class. We normalize both terms, and we obtain the weights \(w\) and \(1-w\) to balance the Binary Cross Entropy.

\subsubsection{Suppressing the loss outside the region of the liver:}

The segmentation of the liver allows us to crop the region of interest in order to segment the lesion. Nevertheless, as we know that the lesion is always inside the liver, we can further benefit from the liver segmentation, deciding not to back-propagate gradients through those pixels that are predicted as non-liver. The benefits of this strategy are twofold: (i) the network is just learning from the pixels that actually can belong to the target class, and (ii) the positive and negative pixels are more balanced, as the number of negative pixels is significantly reduced. Accordingly, the weighting terms of the loss objective just consider the pixels that belong to the liver class. 

\subsubsection{Using context slices to exploit 3D information}
\label{sec:3-slices}
Until this stage, we have been dealing with the data as if each image was independent from the others, but actually we have volumes of images that present spatial coherence. We could benefit from the redundancy among consecutive slices by inputting a volume to the network.  Since we are training from the pre-trained weights of  Imagenet, the network  expects a 3-channel input, so we decided to use these 3 channels to input 3 consecutive slices and segment all of them simultaneously. At test time, we just keep the central slice from the output volume. 

\subsection{Lesion detection network}
\label{sec:detection}
\begin{figure}[h!]
  \centering
  \includegraphics[width=1\textwidth]{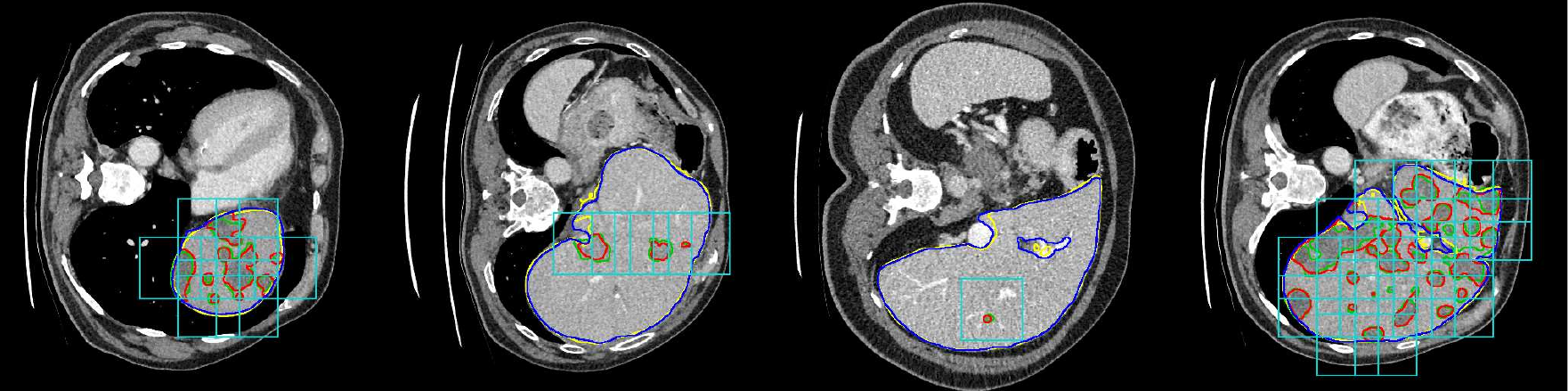}
  \caption[Visualizations of detections]{Results of the lesion detection network. Blue and red lines indicate the liver and lesion ground truth, respectively. Yellow and green lines are the segmentation results for liver and lesion. The light blue bounding boxes are the windows detected as having a lesion. All positive pixels at the output of the segmentation network will be removed if they disagree with the results of lesion detection.}
  \label{Image:detector}
\end{figure}

We observed that our segmentation network lacked the ability to capture a global view of the healthiness of the liver, which is helpful to see if there is a lesion or not, and as a consequence some false positives were triggered in many of the images. A lesion detector acquires a more global insight of the healthiness of a liver tissue compared to the segmentation network, whose final output is pixel-wise and is not forced to take a global decision over a whole liver patch. We will use the detector to localize the lesions and keep those pixels that both the detector and the segmentation network agree that are unhealthy. Figure~\ref{Image:detector} illustrates some examples of how the detector performs.

In order to train the detector, we place bounding boxes in the liver region and then label them as positive or negative.  The condition in order to place a bounding box is that it overlaps at least 25\% with the liver. We use windows of 50$\times$50 pixels, considering a positive observation if there are at least 50 pixels of lesion inside the box. The stride is 50 pixels, and 15 pixels of margin are added to every side of the window to provide additional context, so that each window is finally of size 80x80. The model we used is ResNet-50~\cite{he2016deep} pre-trained on ImageNet, removing its last classification layer and replacing it by a single neuron to detect healthy/unhealthy liver tissues. We use a batch size of 64 equally balanced positive and negative patches. We augment the data by a factor of 8 using flipping and rotation. 

\subsection{Postprocessing}

As a final post-processing step, we add a 3D Fully Connected Conditional Random Field. A 3D-CRF is a statistical modeling technique applied for structured predictions. CRFs model the conditional distribution of the output prediction considering all the input at once. The final labels are assigned given the soft predictions outputted by the segmentation network as a maximum a posteriori inference in a dense CRF. The model considers both the spatial coherence and also the appearance in terms of the intensity values of the input volume. The 3D-CRF is Fully Connected, so it establishes pairwise potentials on all pairs of pixels in the image, maximizing label agreement among similar pixels. We used the implementation of~\cite{christ2017automatic} that uses the 3D-CRF formulation of~\cite{krahenbuhl2011efficient}.

\section{Experimental validation}

Regarding preprocessing of the images from the LiTS dataset, we first clipped the pixels outside the range (-150, 250) of the original images to the minimum/maximum value, as we observed that liver and lesions belong to this limited range. We then performed a min-max normalization of every volume. In Table \ref{Table:test} we show some results in terms of Dice score for the official test set of the LiTS dataset. The first submission we performed was the baseline of a cascaded architecture. For this baseline configuration, the balancing strategy for the loss consists in computing a different balancing term for each CT volume, which is the proportion of positive or negative pixels compared to the total volume. The \textit{Segmentation-only 3-i/o + BP in liver} already includes all the features explained in Section \ref{seg-net}, so it inputs 3 consecutive slices in the network, and limits the samples from which the lesion segmentation network learns to those belonging to the liver. We observe that this configuration yields considerably better results than the baseline. The table also shows the improvements achieved by the detector and the 3D-CRF for the official testing set of the LiTS Challenge. 

\begin{table}[]
\centering
\begin{tabular}{l|l}
                                                            & Dice \\ \hline
Segmentation-only baseline                      & 0.41       \\
Segmentation-only 3-i/o + BP in liver                    & 0.54       \\
Segmentation-only 3-i/o + BP in liver + Detector            & 0.57       \\
Segmentation-only 3-i/o + BP in liver + Detector + 3D-CRF & 0.59      
\bigskip
\end{tabular}
\caption{Results on the LiTS Challenge test (4 different submissions performed on the official Challenge site). \textit{Segmentation-only baseline} refers to the baseline of using a cascaded architecture that first segments the liver in order to later segment its lesions. \textit{Segmentation-only 3-i/o + BP in liver } configuration is the one that inputs 3 consecutive slices in the network, and that just back-propagates through liver pixels when training the network to segment the lesion.}
\label{Table:test}
\end{table}

\section{Conclusions}

In this work we have proposed an algorithm for segmenting the liver and its lesions using two cascaded segmentation networks and a lesion detector. We have studied how to exploit the characteristics of the provided data, using the volume information at the input of the network, and taking advantage of the liver segmentation prediction in order to segment the lesion. 

Our method to segment the lesions of the liver just learns from pixels of the liver, which suggests that limiting the samples from which the algorithm learns to just relevant samples, or difficult ones, is favorable to the problem. This strategy is familiar to using ``Attention'' mechanisms, as there is a location selected (the liver) from which to learn and this improves the learned representations for the lesion. Another advantage of this strategy is that limiting the learning to some samples leverages the imbalance between positive and negative pixels. 

The most important conclusion from this work is the trade-off between fine localization that the segmentation network can achieve, and the generalization that the detector learns. As the output of the segmentation network is pixel-wise, it tends to trigger false positive pixels, since no constraints for more global decisions are imposed. On the other hand, the detector decides if a complete patch is healthy or not, without constraints on the exact shape of the lesion. Having both techniques analyzing the input image yields a better overall result for the LiTS Challenge database. We believe that this is an interesting direction for medical image segmentation pipelines, which typically deal with very small structures, being a detection-aided segmentation pipeline beneficial to localize the target region.

\subsubsection*{Acknowledgments}

This work was partially supported by the Spanish Ministry of Economy and Competitivity under contracts TIN2012-34557 by the BSC-CNS Severo Ochoa program (SEV-2011-00067), and contracts TEC2013-43935-R and TEC2016-75976-R. It has also been supported by grants 2014-SGR-1051 and 2014-SGR-1421 by the Government of Catalonia, and the European Regional Development Fund (ERDF).

\small

\bibliographystyle{plainnat}
\bibliography{references}

\end{document}